\documentclass[letterpaper, 10 pt, conference]{ieeeconf}  

\IEEEoverridecommandlockouts                              
\usepackage{amsmath,amssymb,amsfonts, mathtools}
\usepackage{graphicx}
\usepackage{bm}
\usepackage{placeins}
\usepackage{bm}
\usepackage{cite}
\let\labelindent\relax
\usepackage[inline]{enumitem}
\usepackage{multicol}
\usepackage{tikz}
\usepackage{caption}
\usepackage{subcaption}
\usepackage{soul}
\usepackage{wrapfig}

\usepackage{enumitem}

\title{\LARGE \bf Benchmark for Skill Learning from Demonstration: Impact of User Experience, Task Complexity, and Start Configuration on Performance}

\author{M. Asif Rana$^1$, Daphne Chen$^1$, S. Reza Ahmadzadeh$^2$, Jacob Williams$^1$, Vivian Chu$^1$, and Sonia Chernova$^1$
	\thanks{$^1$ Georgia Inst. of Technology, Atlanta, GA. Email: {\tt\small \{asif.rana,daphne.chen,chernova\}@gatech.edu}}%
	\thanks{$^2$ University of Massachusetts Lowell, Lowell, MA. Email: {\tt\small reza\_ahmadzadeh@uml.edu}}}

\newenvironment{flushitemize}{%
\begin{list}{$\bullet$}
   {\setlength{\leftmargin}{15pt}}%
    \setlength{\labelwidth}{20pt}
    \setlength{\itemindent}{0pt}
    \setlength{\labelsep}{0.5em}
 \setlength{\itemsep}{1pt}
 \setlength{\parskip}{0pt}
 \setlength{\parsep}{0pt}}
 {\end{list}}

\begin{document}

\maketitle
\thispagestyle{empty}
\pagestyle{empty}

\begin{abstract}
In this work, we contribute a large-scale study benchmarking the performance of multiple motion-based learning from demonstration approaches. Given the number and diversity of existing methods, it is critical that comprehensive empirical studies be performed comparing the relative strengths of these learning techniques. In particular, we evaluate four different approaches based on properties an end user may desire for real-world tasks. To perform this evaluation, we collected data from nine participants, across four different manipulation tasks with varying starting conditions.  The resulting demonstrations were used to train 180 task models and evaluated on 720 task reproductions on a physical robot. Our results detail how i) complexity of the task, ii) the expertise of the human demonstrator, and iii) the starting configuration of the robot affect task performance. The collected dataset of demonstrations, robot executions, and evaluations are being made publicly available. Research insights and guidelines are also provided to guide future research and deployment choices about these approaches.
\end{abstract}

\section{Introduction}
\label{sec:intro}

Robots must have the capability to continuously learn new skills in order to accomplish a variety of tasks in dynamic and unstructured environments. Learning from demonstration (LfD)~\cite{argall2009survey} aims to enable robots to continuously acquire such skills from human interaction without the need for manual programming. 
 
In this paper we focus on learning robot motions from human demonstrations. For learning a desired motion-based skill, a model is typically trained over multiple trajectory demonstrations collected from a human end user. During reproduction, either in a previously seen or novel scenario, the learned model is queried to generate new executable trajectories. A scenario generally includes starting position from where the end user desires the robot to execute the task at hand.

From the perspective of an end user, there are multiple desirable properties that a motion-based skill learning approach should have, including the ability to:
\begin{enumerate}[label=A.{\arabic*}]
	\item learn skills from demonstrations provided by end users irrespective of all experience levels, with minimal information overload on the user,
	\item learn a variety of skills, which may differ in the level of complexity, and
	\item reproduce a learned skill in scenarios similar to or different from those encountered while collecting the demonstrations.
\end{enumerate}

\begin{figure}
	\centering
	\includegraphics[trim={0cm 0cm 0cm 0cm}, clip, width=0.95\columnwidth]{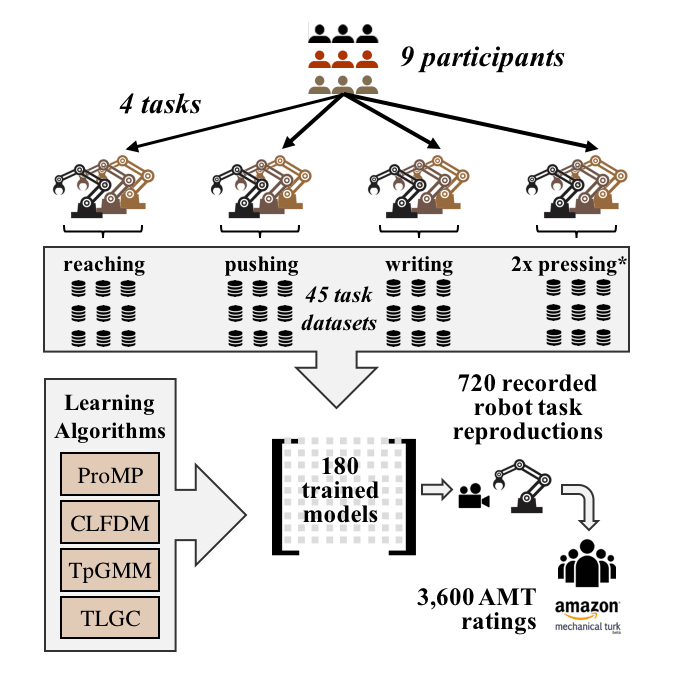}
	\caption{\small{Overview of experimental design. 9 participants executed 4 tasks on the robot. These demonstrations were used to train 4 trajectory learning algorithms, resulting in models that were executed on the robot to reproduce each original task. The reproductions were evaluated via crowdsourcing on Amazon Mechanical Turk.}}
	\label{fig:overview}
\end{figure}

Given the number and diversity of existing motion-based LfD approaches, it is critical that comprehensive empirical studies be performed to compare the relative strengths of these learning techniques. The majority of techniques can be broadly categorized into one of four categories based on choice of model representation: statistical approaches~\cite{calinon2007learning, calinon2016tutorial}, dynamical systems~\cite{khansari2011learning, khansari2014learning, ijspeert2013dynamical, chaandar2018learning}, geometric techniques~\cite{nierhoff2016spatial,meirovitch2016geometrical,ahmadzadeh2017generalized}, or probabilistic inference~\cite{rana2017towards,paraschos2013probabilistic,huang2017kernelized,schneider2010robot}. However, comprehensive work which evaluates these approaches based on the criteria mentioned earlier does not exist to date. 

In this work, we evaluate the performance of multiple motion-based skill learning approaches and examine how the i) complexity of the task, ii) expertise level of the human demonstrator, and iii) starting configuration of the robot affect performance of each technique. For our evaluation, we compared four algorithms, namely TpGMM~\cite{calinon2016tutorial}, CLFDM~\cite{khansari2014learning}, TLGC~\cite{ahmadzadeh2017generalized}, and ProMP~\cite{paraschos2013probabilistic} – one from each aforementioned category. Our selection targets techniques that  are  most  well-known,  commonly  used,  or are most mature based on incremental improvement on prior work.
To perform this evaluation, we collected data from nine participants across four different manipulation tasks with varying starting conditions. The resulting demonstrations were used to train 180 task models. Each of the resulting models was then executed on a Rethink Sawyer robot, resulting in 720 videos of robot task reproductions. Finally, we obtained 3600 Amazon Mechanical Turk ratings to evaluate the robot's performance in the videos. Fig. \ref{fig:overview} provides an overview of our experimental procedure. Additionally, we present an evaluation based on quantitative error metrics obtained by assessing the similarity between the reproduced trajectories and the demonstrations. The full dataset of demonstrations, videos of executions, and accompanying evaluations have been made publicly available to aid future benchmarking efforts\footnote{https://sites.google.com/view/rail-lfd}.

Our results show that the performance of the skill learning approaches --- irrespective of their underlying representation --- is generally predictable when the new starting condition is closer to the starting position of demonstrations. However, as the generalization scenario differs from the demonstrations, the consistency of an approach's performance across generalization scenarios is highly dependent on the task constraints. Furthermore, we also find that the performance of a given skill learning method is correlated with the experience level of the human providing demonstrations. Lastly, we found that commonly used performance evaluation metrics such as mean squared error are not always able to correctly predict the generalization performance of an approach. 

The authors intend for this work to be used by those who study LfD by acting as a reference for experimental design, evaluation metrics, and general best practices.

\section{Related Work}

In this section, we present an overview of motion-based LfD and describe the techniques examined in our study.

\subsection{Overview of Motion-Based LfD}
There exist several approaches aimed at learning motion-level skills from human demonstrations. Among them are reactive approaches, often based on learning dynamical systems~\cite{khansari2011learning,khansari2014learning,perrin2016fast,ravichandar2017learning,neumann2015learning}, while others are based on learning time-parametrized representations of motions~\cite{paraschos2013probabilistic,calinon2007learning, rana2017towards}. Within these categories are further subcategories divided on the choice of skill representation. In literature however, these approaches are usually tested in isolation by experts for a specific set of tasks. While the relative advantages and disadvantages of the commonly-used approaches might be known within the LfD community, there do not exist comprehensive guidelines for non-experts outside the community to assist in using these methods. Comprehensive surveys on LfD~\cite{osa2018algorithmic, argall2009survey, kroemer2019review, hussein2017imitation} do exist, but they mainly focus on summarizing existing LfD approaches, proposing taxonomy, and reporting challenges associated with employing LfD approaches in practice. There is a need to supplement these surveys by comparing and evaluating LfD approaches across several variables that can be encountered in the real world.

Prior work by Lemme \textit{et al.} contributed a valuable benchmarking framework to evaluate the performance of point-to-point reaching motion generation approaches on a 2D handwriting dataset~\cite{lemme2015open}. Their study evaluates the algorithms' generalization ability in simulation and presents performance metrics on a small scale. Our study is more comprehensive: it covers multiple tasks, incorporates diverse constraints and variables, and is performed on a physical robot. 

To our knowledge, no prior benchmarking study exists that independently evaluates a wide range of task execution conditions.  Additionally, no prior studies report human ratings of task performance. 

\makeatletter
\setlength{\@fptop}{0pt}
\makeatother
\begin{figure*}[t!]
  \begin{subfigure}[b]{0.24\linewidth}
    \includegraphics[width=0.93\columnwidth]{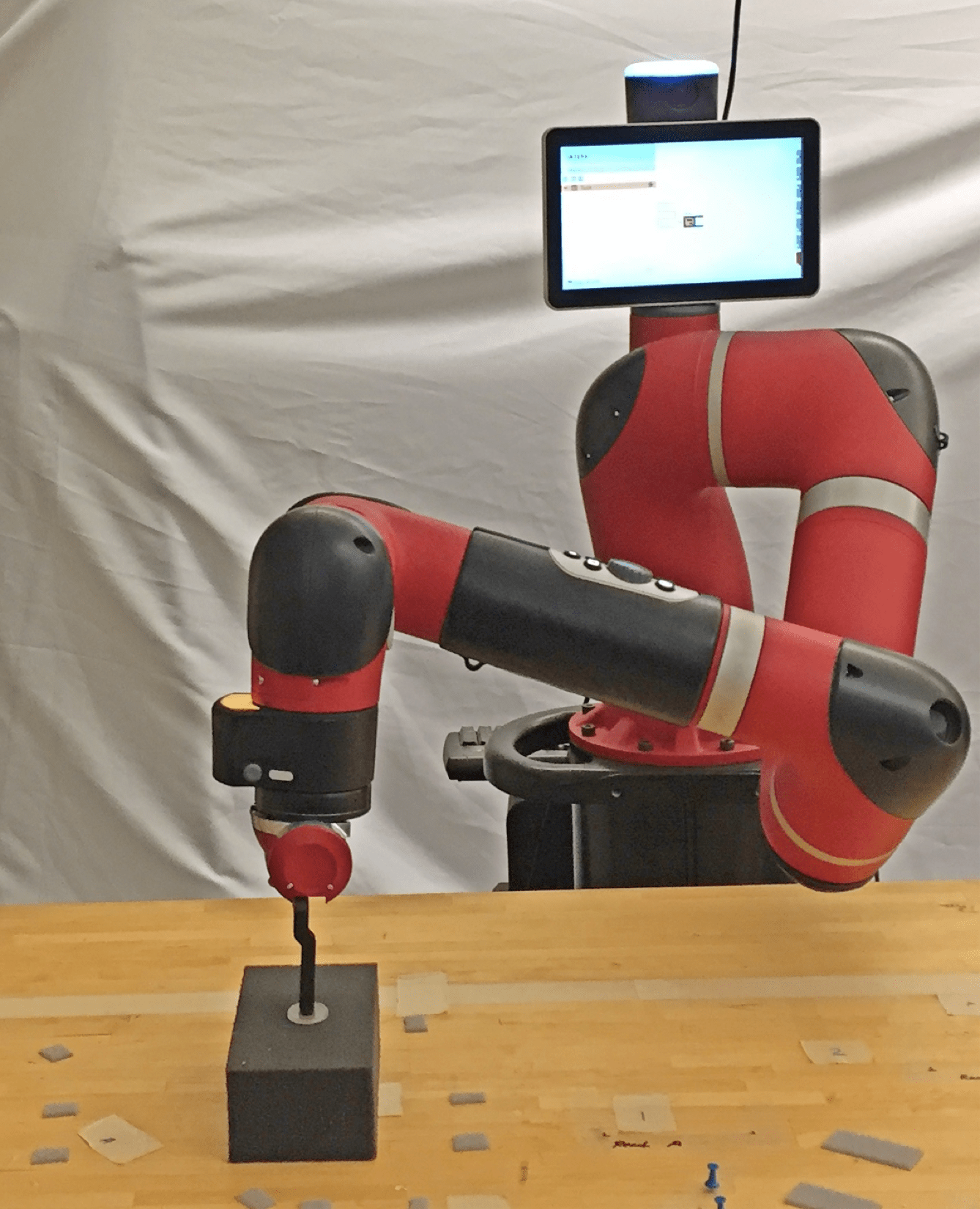}
    \caption{}
  \end{subfigure}
  \hfill
  \begin{subfigure}[b]{0.24\linewidth}
    \includegraphics[width=\columnwidth]{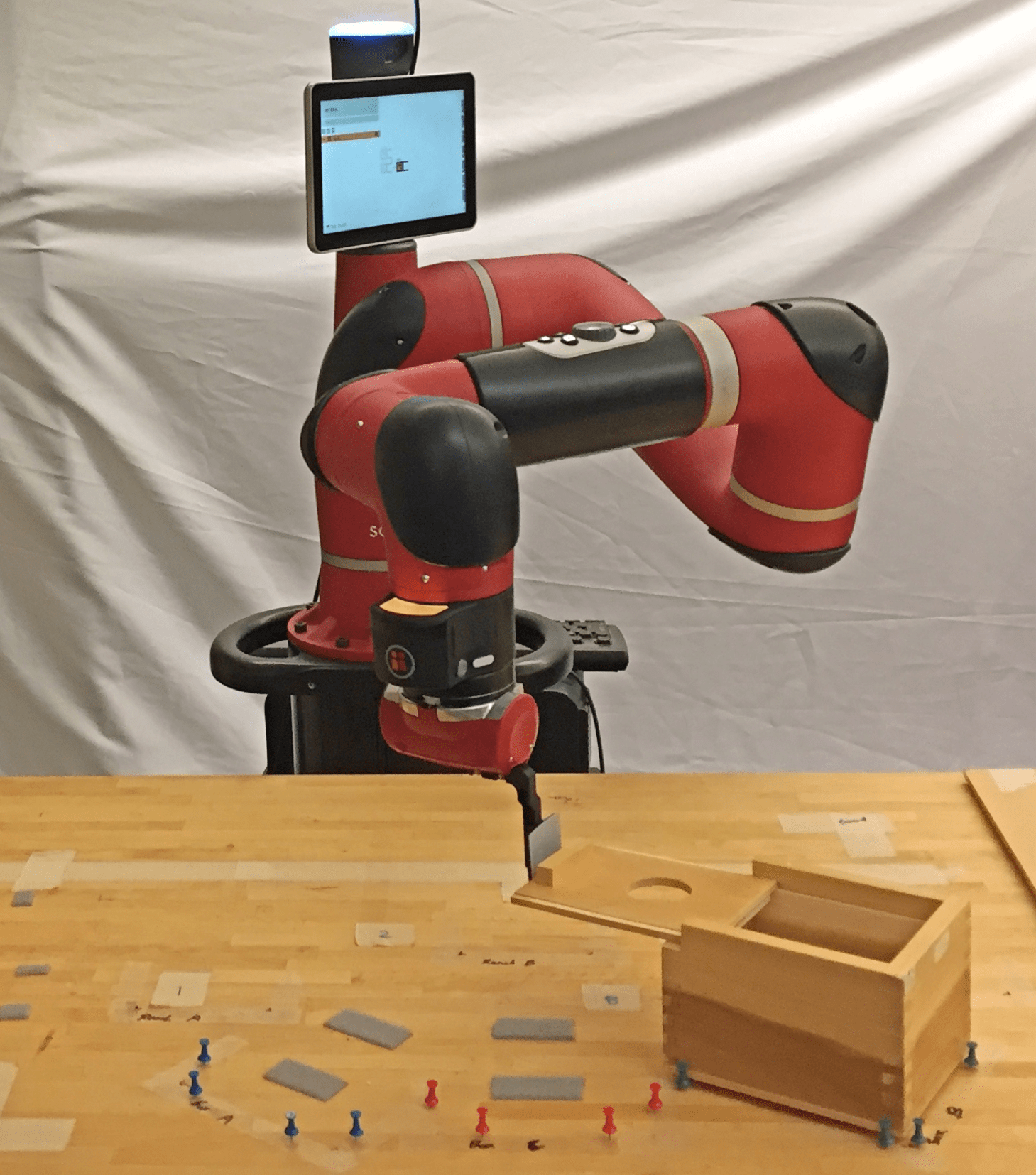}
    \caption{}
  \end{subfigure}
  \hfill
  \begin{subfigure}[b]{0.24\linewidth}
    \includegraphics[width=0.92\columnwidth]{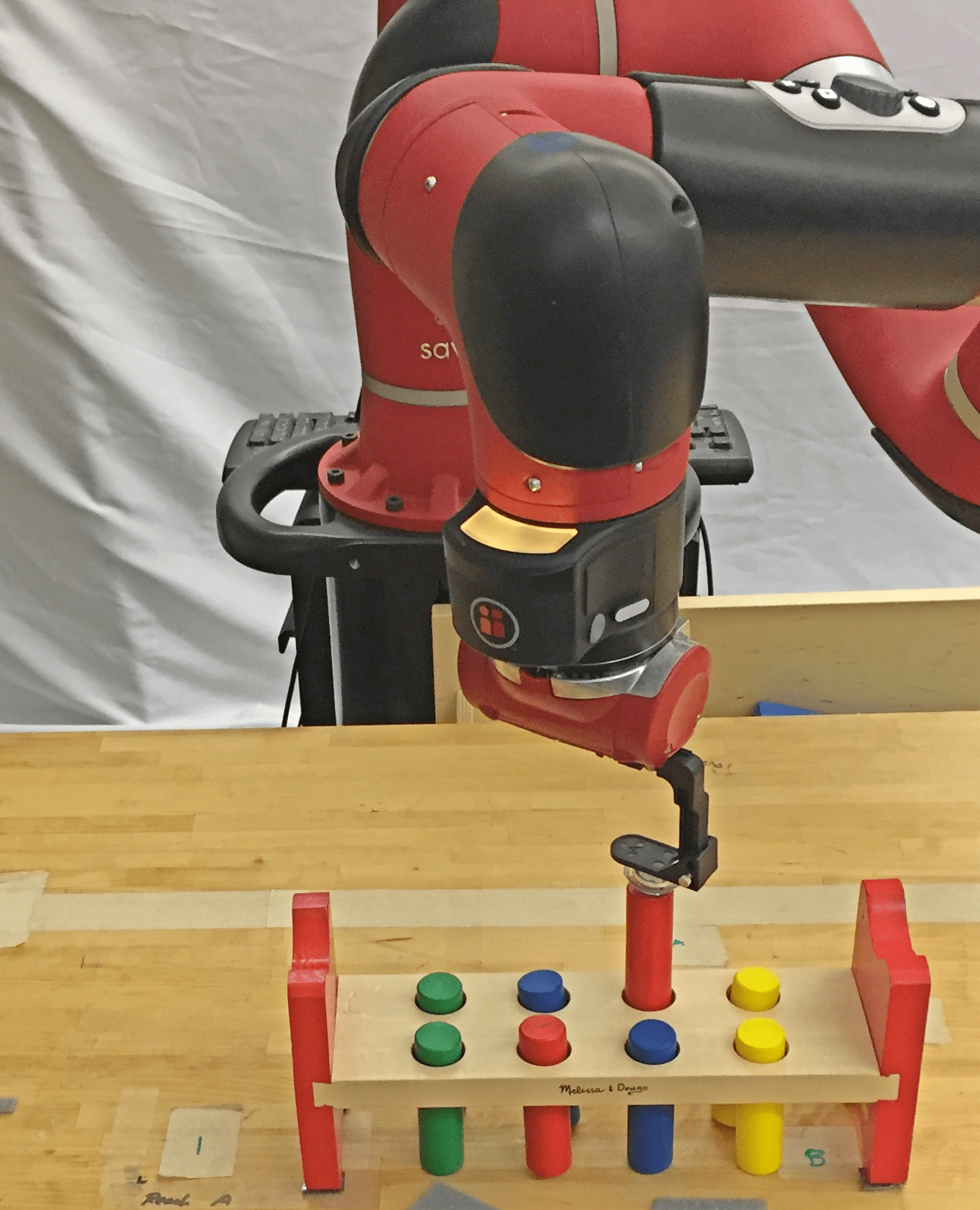}
    \caption{}
  \end{subfigure}
  \hfill
  \begin{subfigure}[b]{0.24\linewidth}
    \includegraphics[width=\columnwidth]{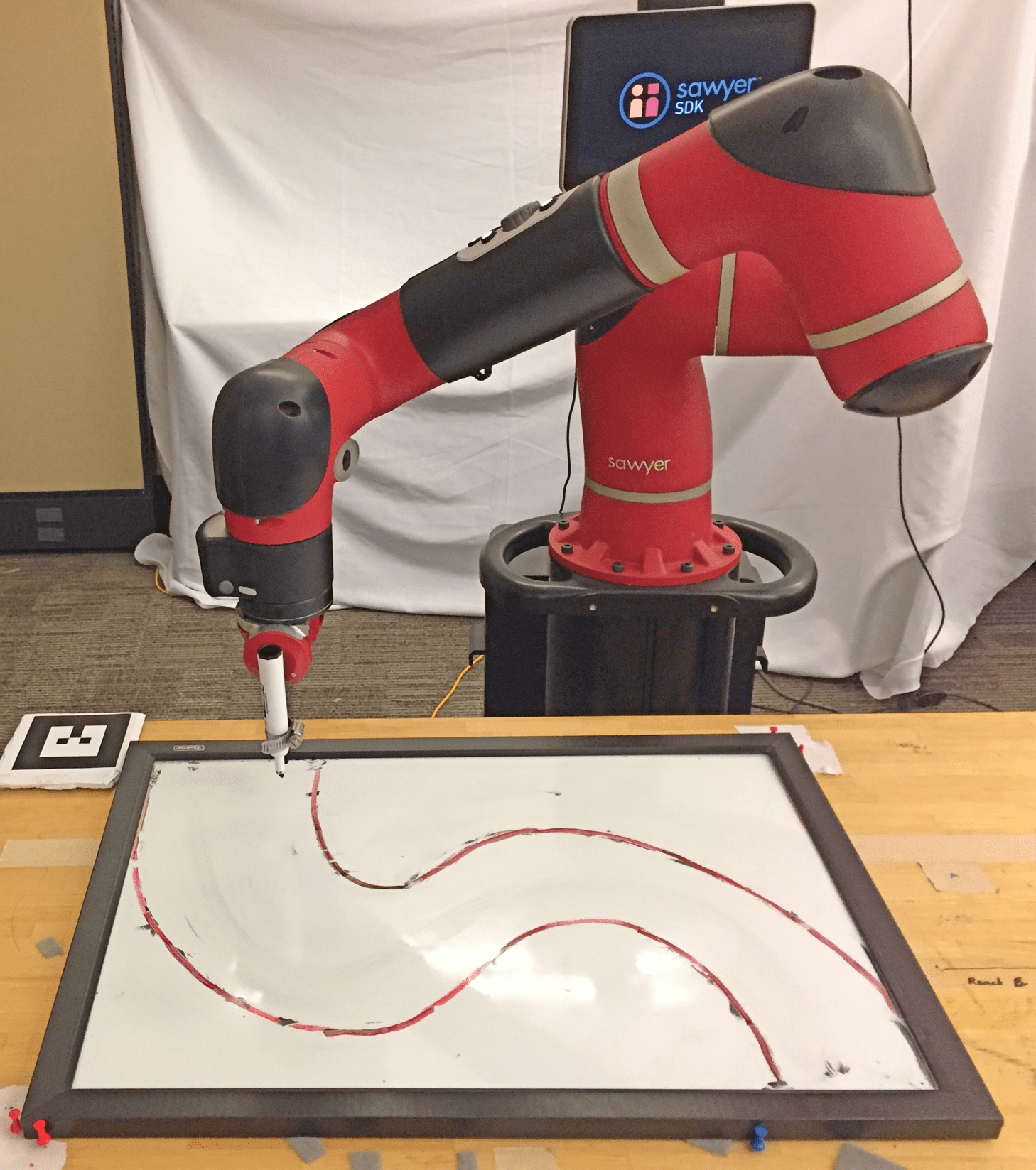}
    \caption{}
  \end{subfigure}
   \begin{subfigure}[b]{0.24\linewidth}
    \includegraphics[width=\columnwidth]{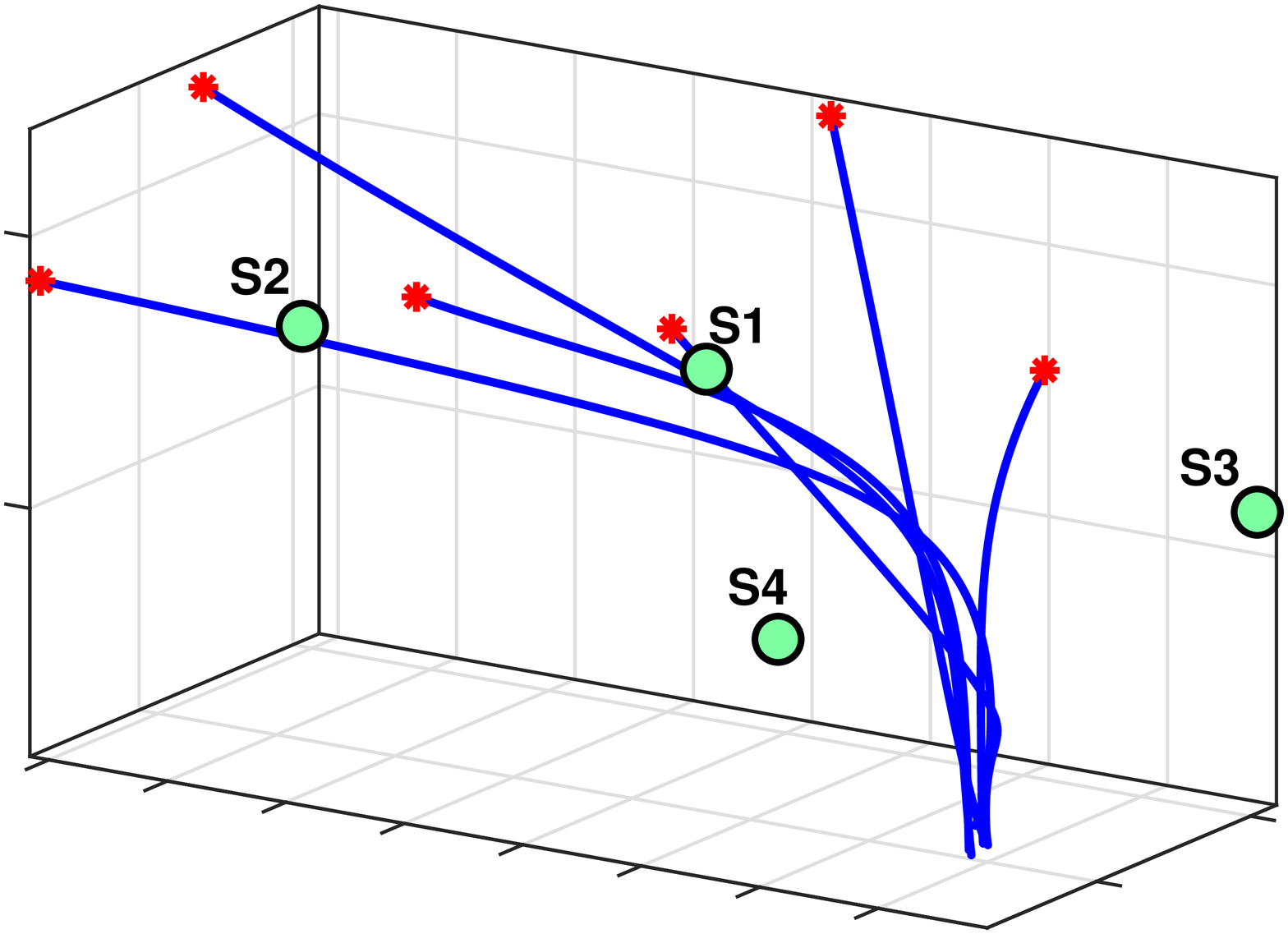}
  \end{subfigure}
  \hfill
  \begin{subfigure}[b]{0.24\linewidth}
    \includegraphics[width=0.82\columnwidth]{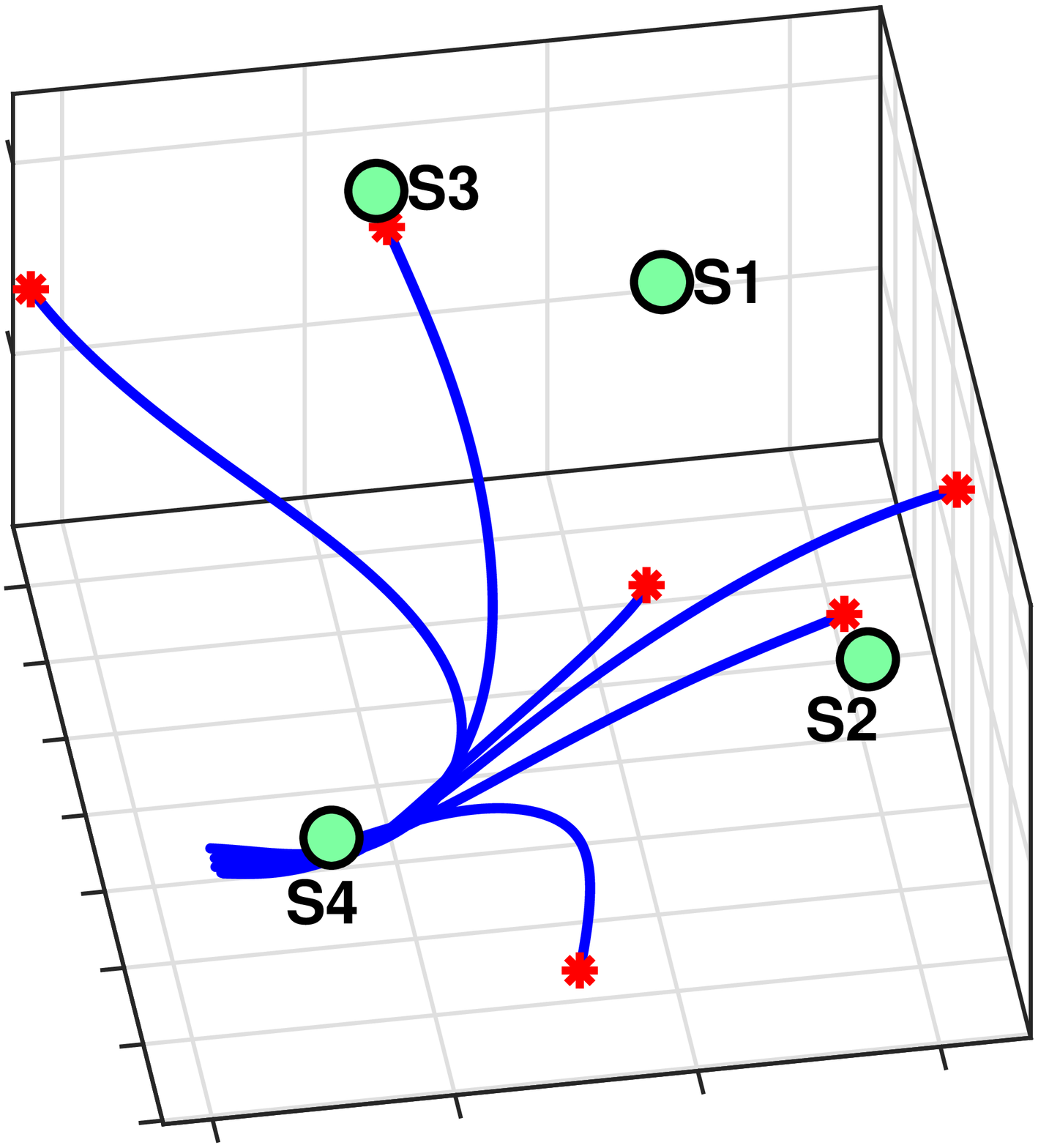}
  \end{subfigure}
  \hfill
  \begin{subfigure}[b]{0.24\linewidth}
    \includegraphics[width=0.96\columnwidth]{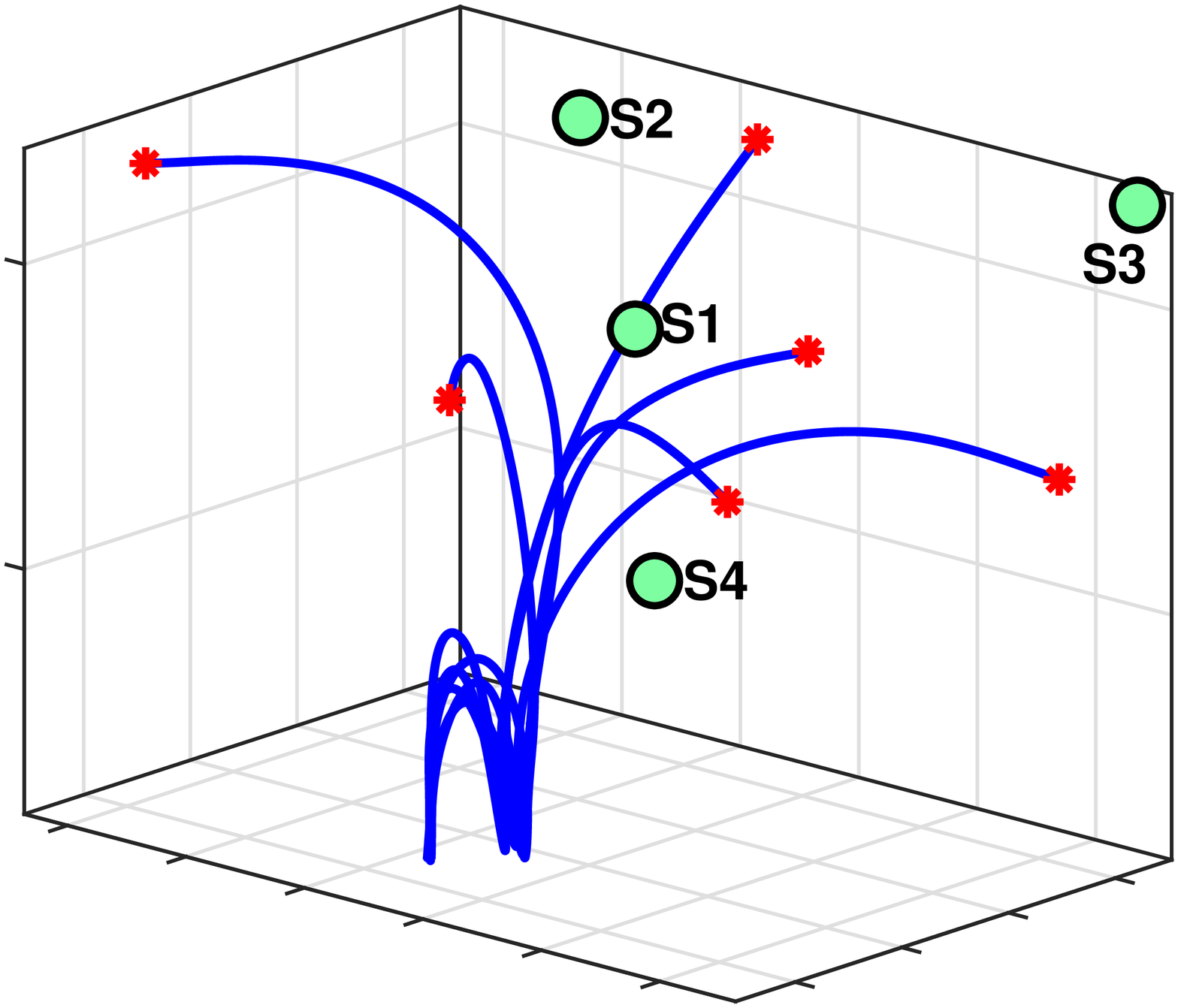}
  \end{subfigure}
  \hfill
  \begin{subfigure}[b]{0.24\linewidth}
    \includegraphics[width=\columnwidth]{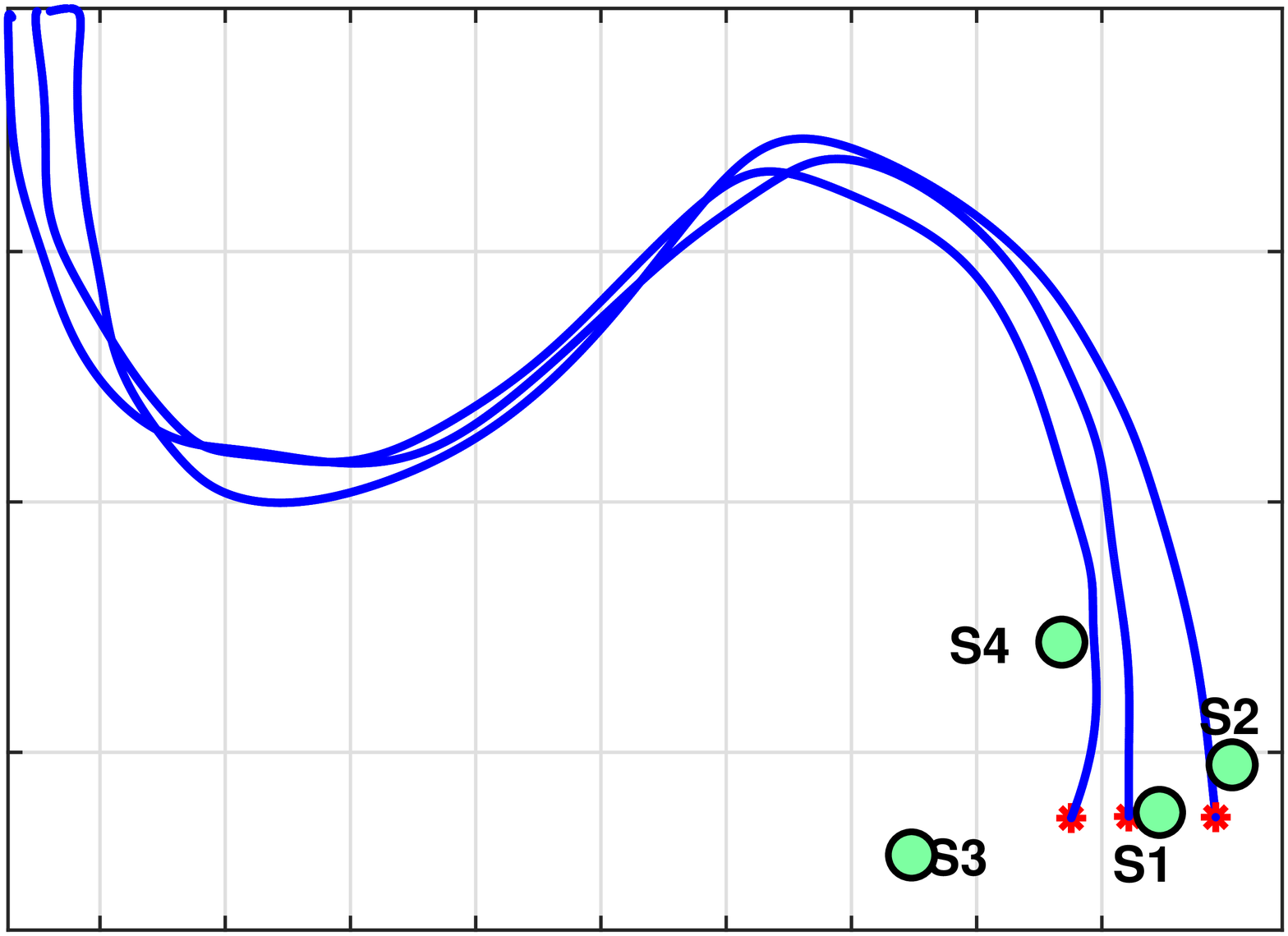}
  \end{subfigure}
  \caption{\small{From left to right, the top row shows the robot executing the benchmarking tasks: \emph{reaching}, \emph{pushing}, \emph{pressing}, and \emph{writing}; the bottom row shows a visualization of an example dataset collected for each task. The red dots show the different starting points and the blue lines show the trajectories. The green circles illustrate the \emph{new} initial positions selected for evaluating skill generalization.}}
\label{fig:task-executions}
\end{figure*}

\subsection{Techniques Selected for Comparison}
The algorithmic techniques evaluated in this work were chosen to represent each of the four categories of model representations mentioned earlier. 

While many other LfD approaches exist, we have selected these four approaches because they are well-known, commonly used, or are most mature based on incremental improvement on prior work. 

Below, we provide a brief description of each method; please refer to the references for full details. 

\noindent \textit{\textbf{CLFDM}}\cite{khansari2014learning} -- An approach which learns a stable dynamical system from demonstrations. Specifically, CLFDM fits a dynamics model of the form $\dot{\bm{x}}_t = \bm{f}(\bm{x}_t)$. 
The dynamical system is composed of two components $\bm{f}(\bm{x}_t) = \hat{\bm{f}}(\bm{x}_t) + \bm{u}(\bm{x}_t)$, where  $\hat{\bm{f}}(\bm{x}_t)$ is an unconstrained regression model and $\bm{u}(\bm{x}_t)$ is a stabilizing controller. It is assumed here that the final positions of the demonstrated motions are centered at a single goal location, and hence the stabilizing controller ensures that trajectory roll-outs always converge to this goal.

\noindent \textit{\textbf{ProMP}}\cite{paraschos2013probabilistic} -- A probabilistic approach which learns a time-dependent stochastic controller from demonstrations. ProMP finds a controller $\bm{u}_t = \bm{f}_t(\bm{x}_t, \dot{\bm{x}}_t) + \bm{\epsilon}_t$, where  $\bm{f}_t(\cdot)$ is a time-varying feedback control law and $\bm{\epsilon}_t$ represents a time-varying Gaussian control noise. By rolling out the system using this stochastic controller, this approach generates a distribution of trajectories, the mean of which is executed. Additional constraints (e.g. via-points) can be added to the trajectory distribution to modulate the executable trajectory. 

\noindent \textit{\textbf{TLGC}}\cite{ahmadzadeh2017generalized} -- A geometric approach which explicitly encodes the geometric features of the demonstrations. This approach fits a generalized cylinder to the demonstrations. Given a new initial position, a ratio $\rho_0$ is found for the distance from the initial position to the center of the cylinder and the distance to the closest point on the boundary of the cylinder. A new trajectory is found by maintaining this ratio while traversing the arc length of the generalized cylinder. This approach uses the Laplacian trajectory editing technique for generalization and can reproduce trajectories using multiple reproduction strategies\cite{ahmadzadeh2018trajectory}.

\noindent \textit{\textbf{TpGMM}}~\cite{calinon2016tutorial} -- An approach which encodes the statistical features of the demonstrations. TpGMM finds a time-dependent mean $\bm{\mu}_t$ and variance $\bm{\Sigma}_t$ of the demonstrations. New trajectories are generated by solving an LQR problem which seeks to find the smallest sequence of controls that penalize deviations from $\bm{\mu}_t$ weighted by the inverse of $\bm{\Sigma}_t$. In effect, TpGMM carries out \emph{minimum intervention control} whereby it tracks a reference trajectory with variable stiffness. Additionally, TpGMM allows encoding demonstrations in which multiple objects might be relevant and the reproduction has to adapt to changes in their locations.

\section{Experimental Design}

This section provides an overview of our experimental design process, including choice of tasks, human participant selection, as well as the methodology for data recording, model evaluation, and Amazon Mechanical Turk (AMT) rating.  Fig.~\ref{fig:overview} presents a summary of the full experimental process.

\subsection{Robot Tasks}\label{sec:RobotTasks}
We selected four tasks (Fig. \ref{fig:task-executions}) each of which contains unique properties representing different level of position and motion constraint complexity. Human demonstrator ability was kept in mind such that users with minimal experience could demonstrate the task on the robot.

\begin{flushitemize}
	\item \emph{Reaching} - Move toward and touch the circle on the gray block (Fig. \ref{fig:task-executions}a). This task poses a hard constraint on the end position. 
	
	\item \emph{Pushing} - Push the box lid closed (Fig. \ref{fig:task-executions}b). Comparing to the previous task, this task is constrained in the direction of motion towards the end. The position constraint for the endpoint is not as hard as in the \emph{reaching} task.
	
	\item \emph{Pressing} - Push down peg $\#1$ and then peg $\#2$ (Fig. \ref{fig:task-executions}c). Compared to \emph{pushing}, this task is more constrained in both the direction of motion as well as end-positions.
	
	\item \emph{Writing} - Draw an S-shaped curve on the whiteboard (Fig. \ref{fig:task-executions}d). Compared to other tasks, this task requires a harder constraint on the direction of motion to follow the curvature of the shape.
\end{flushitemize}

\subsection{Participant Selection}\label{sec:Participant}

The implicit characteristics of human-provided demonstrations affect the performance of the LfD approaches significantly~\cite{argall2009survey}. Therefore, we chose to include demonstrators with different experience levels in our experiments.

We recruited nine participants with different levels of robotics experience from the Computer Science and Engineering community at Georgia Tech. Three participants with \textit{Low} experience had no prior interaction with any type of robot. Three participants with \textit{Medium} experience had worked with robots but had no experience in robot manipulation, and particularly no experience in kinesthetic teaching. Three participants with \textit{High} experience had previously used motion-based LfD methods through kinesthetic teaching.

\vspace{-2mm}
\subsection{Data Recording}\label{sec:Data}

Data collection with participants followed an IRB-approved human subjects study protocol and participants were compensated with a \$10 gift-card.  Upon arrival, participants were briefed about the goals of the study and taught to interact with the robot using a practice task (i.e., pushing a toy car across the table using the robot's end-effector). 

Participants received written instructions that included a verbal description and photos of the goals of each task\footnote{Example: Fig. \ref{fig:task-executions}(a) accompanied by the instruction, ``The robot finger-tip should touch the small circle on the gray block".}. This ensured the consistency of the guidelines across all participants and evaluators. During the recording, the robot was first initialized to a pre-set starting configuration put in gravity-compensation mode. The participant kinesthetically guided the robot to accomplish the task. Finally, the recording was stopped at the participant's command. In order to assess the quality of the demonstrations, we provided the participants with a visualization of the recorded trajectory in ROS RViz. Participants were allowed to perform multiple executions of the task until they were satisfied with the quality of the data; we kept only the final execution.  In total, the participants provided three demonstrations for the \emph{writing} task with three different starting positions. For the remaining tasks, six demonstrations (3 starting positions $\times$ 2 object locations) per participant were collected. This resulted a total of 21 demonstrations per participant. Fig. \ref{fig:task-executions} (\emph{bottom}) shows an example set of demonstrations transformed such that the origin is at the target object location.

\subsection{Model Evaluation}\label{sec:Model}
From the collected demonstrations, we constructed 45 task datasets. Each dataset includes all demonstrations of a specific task (four tasks) performed by a specific participant (nine participants). Note that each participant was asked to demonstrate the \emph{pressing} task twice each time under a different condition (see Section \ref{sec:motion-seg} for more detail.), and as a result $9$ participants $\times 5$ tasks $= 45$ datasets.

Each of our four algorithms was then trained on each of the 45 datasets, resulting in 180 task models (one per participant-task-algorithm combination). For evaluation, we executed each of the 180 models under four different starting conditions on a Sawyer robot, resulting in 720 video recordings of robot task executions over the four tasks. To obtain a final evaluation of the robot's performance in each of the videos, we employed five AMT ~\cite{buhrmester2011amazon} workers to evaluate the quality of each video, resulting in approximately 3600 performance ratings.

\subsection{Amazon Mechanical Turk Evaluation} \label{sec:AMT}

To ensure that AMT workers evaluating the robot had a consistent understanding of the task goals, workers were shown the same set of instructions as those given to the study participants (i.e., task demonstrators). For each video of the robot's task execution, AMT workers were  asked to answer the following questions:

\noindent\fbox{%
    \parbox{\columnwidth}{%
\medskip

	\noindent Q1. Please rate the extent to which you agree with the statement: \emph{``The robot efficiently and safely completed the goal(s) of the task." } (Strongly agree; Agree; Disagree; Strongly disagree).

\smallskip

	\noindent Q2. Please also specify which of the following contributed to your rating in the previous question. (Check all that apply)
	\begin{flushitemize}
		\item The robot failed to achieve the goals of the task (incomplete).
		\item The robot performed unnecessary motion (inefficient).
		\item The robot acted in an unsafe manner (unsafe).
	\end{flushitemize}
}
}
\medskip

Each video was evaluated by five AMT workers and an overall \textit{rating} per video/execution was calculated by taking the median of the responses to the first question. To get a quantitative measure of the evaluator rating, we mapped the answers to numerical values: Strongly agree $= 3$, Agree $=2$, Disagree $=1$, and Strongly disagree $=0$. We consider a task reproduction to be acceptable to the evaluators if the rating is 2 or above. Answers to the second question were only considered if the participant selected a rating below ``Strongly agree'' in response to the first question.

The selected keywords, \emph{incomplete}, \emph{inefficient}, and \emph{unsafe}, are suitable to define the characteristics of the task execution quality from an end user's point of view. Our reasoning is that a robot that cannot complete a task efficiently can impose great burden on the user, and a successful human-robot team requires a  smooth and predictable task execution. 

 \begin{figure*}
 \begin{subfigure}[b]{1\linewidth}
    \centering
    \includegraphics[trim={0cm 0cm 0cm 0cm}, clip, width=0.5\hsize]{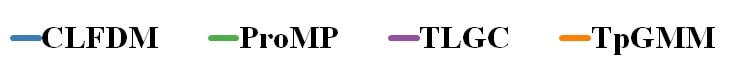}
    \label{fig:legend}
\end{subfigure}
  \begin{subfigure}[b]{0.25\linewidth}
    \includegraphics[trim={2cm 2.3cm 2cm 2.6cm}, clip, width=1.08\linewidth]{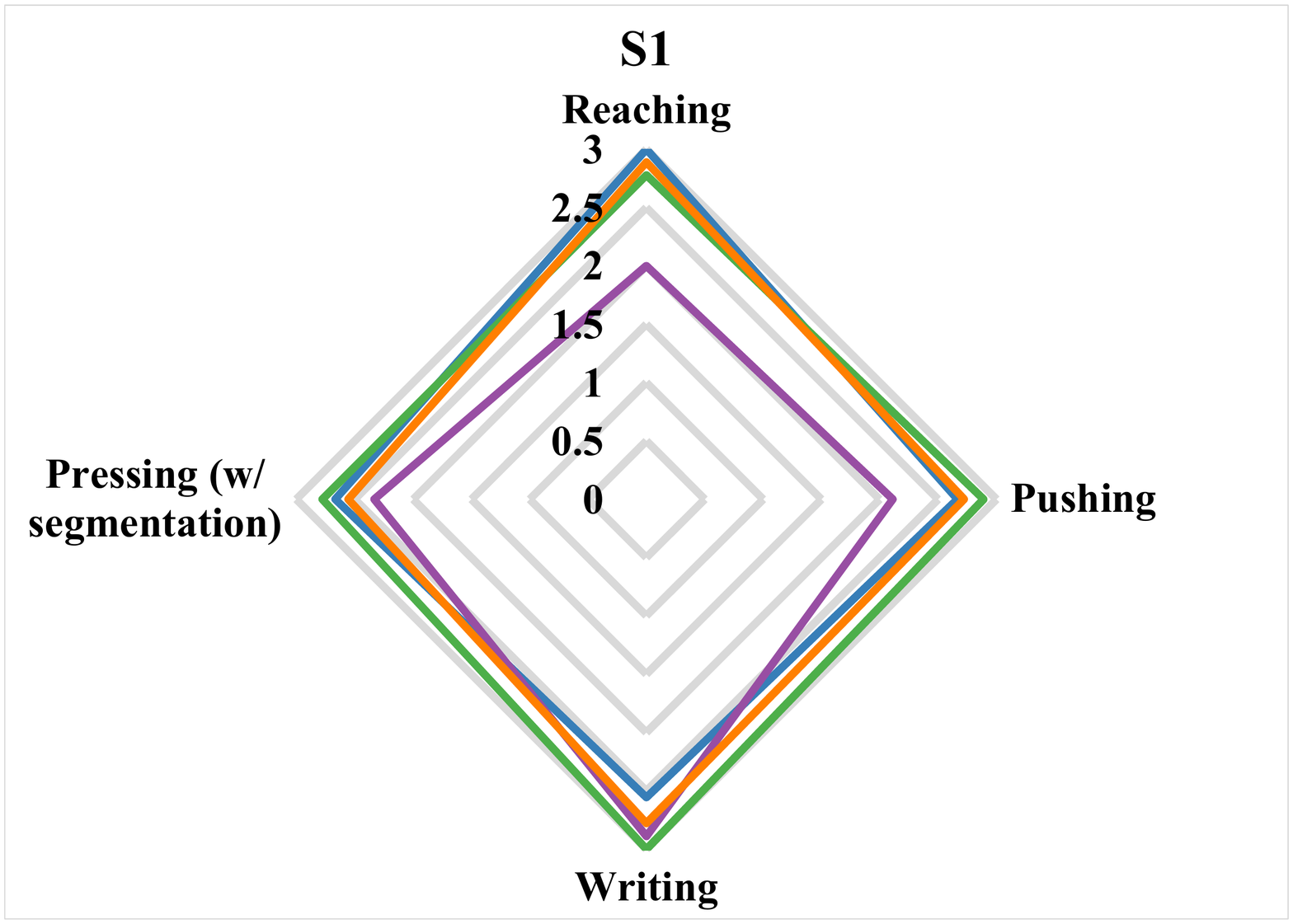}
    \caption{}
    \label{fig:starting-pos-s1}
  \end{subfigure}
 \hspace{-0.2cm}
  \begin{subfigure}[b]{0.25\linewidth}
    \includegraphics[trim={2cm 2.3cm 2cm 2.6cm}, clip, width=1.08\linewidth]{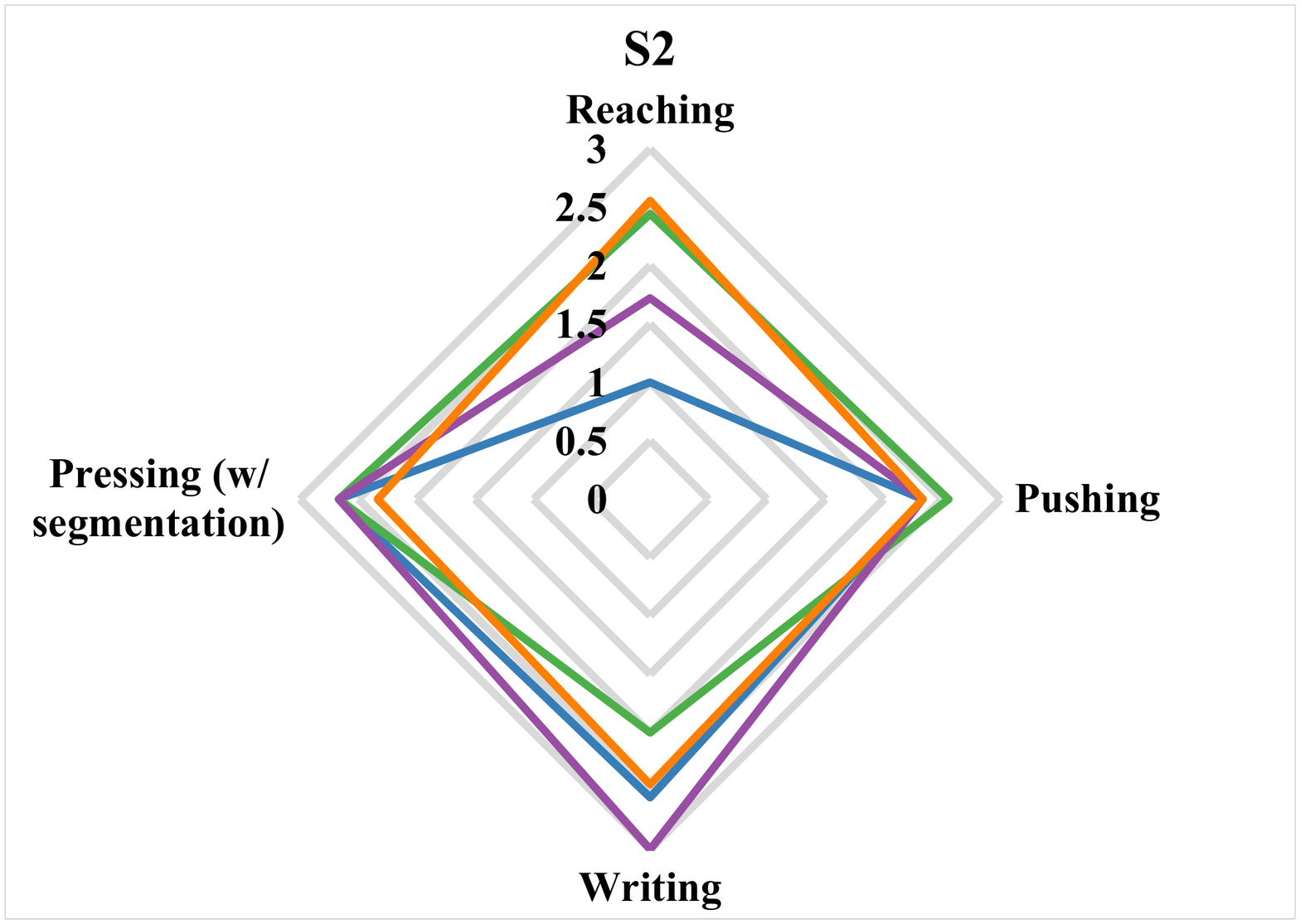}
    \caption{}
    \label{fig:starting-pos-s2}
  \end{subfigure}
  \hspace{-0.2cm}
  \begin{subfigure}[b]{0.25\linewidth}
    \includegraphics[trim={2cm 2.5cm 2cm 2.8cm}, clip, width=1.08\linewidth]{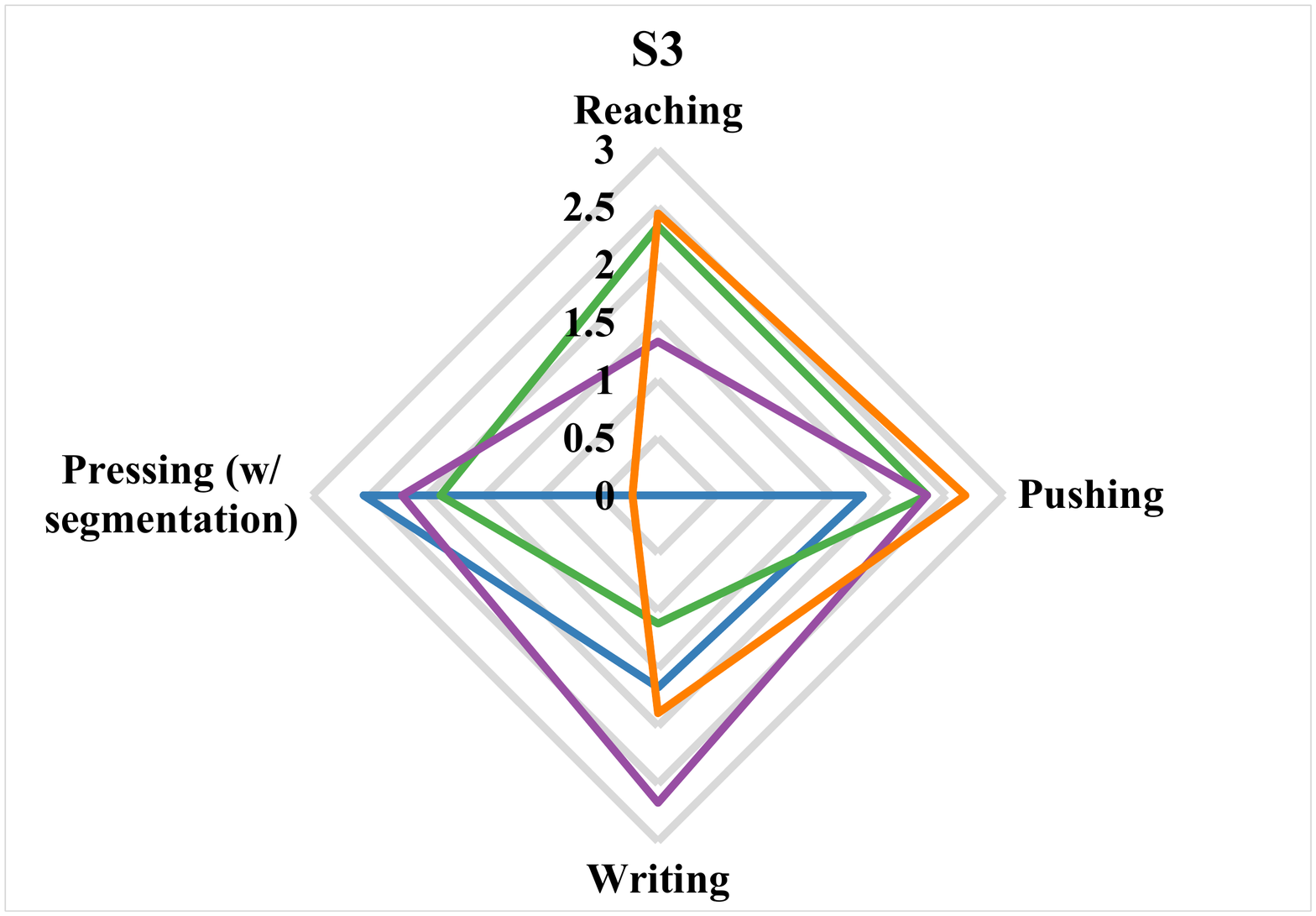}
    \caption{}
    \label{fig:starting-pos-s3}
  \end{subfigure}
  \hspace{-0.2cm}
  \begin{subfigure}[b]{0.25\linewidth}
    \includegraphics[trim={2cm 2.5cm 2cm 2.8cm}, clip, width=1.08\linewidth]{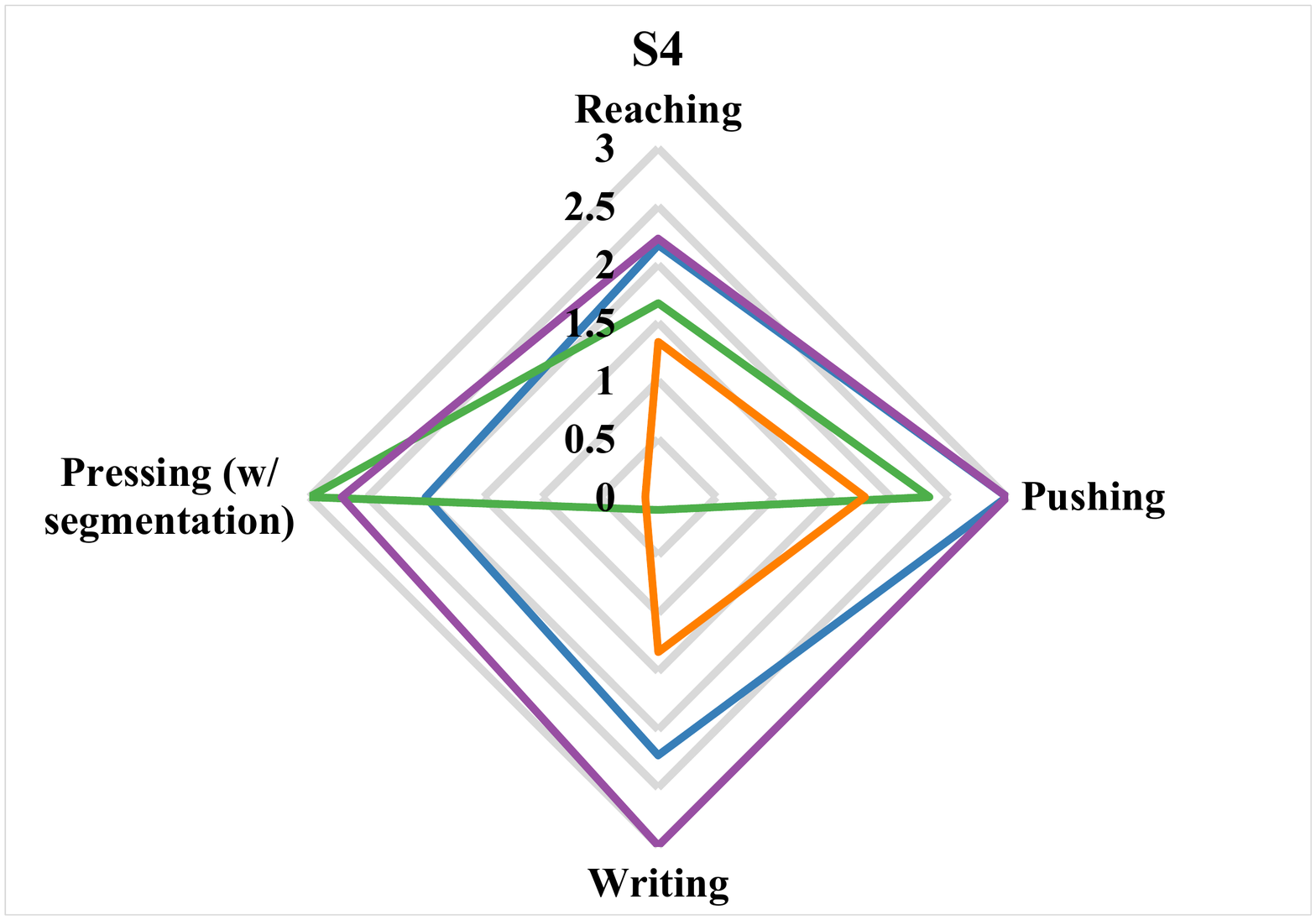}
    \caption{}
    \label{fig:starting-pos-s4}
  \end{subfigure}
\caption{\small{Radar plots of average user rating. The major axis show average ratings, while the corners denote the different tasks.}}
\label{fig:starting-pos}
\end{figure*}

\begin{figure*}
\begin{multicols}{3}
    \includegraphics[trim={2cm 6cm 2cm 6cm}, clip, width=0.95\columnwidth]{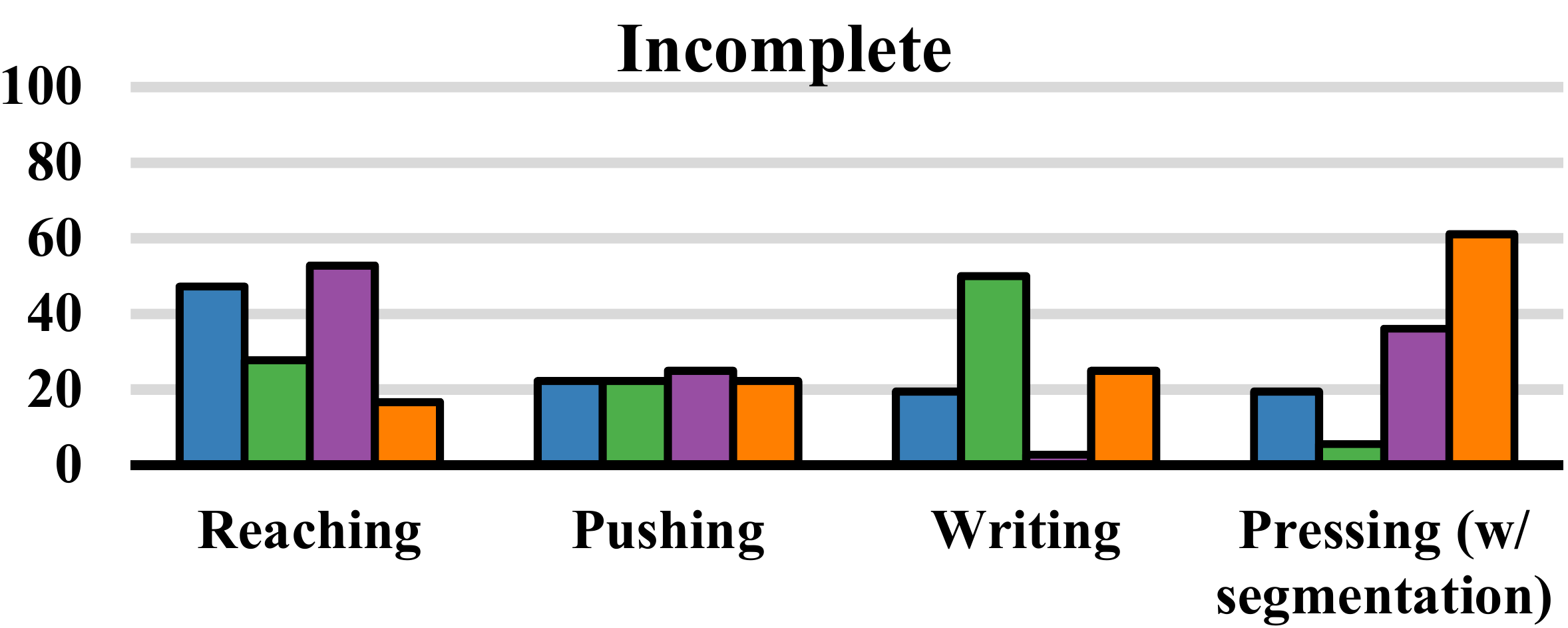}\par   
    \includegraphics[trim={2cm 6cm 2cm 6cm}, clip, width=0.95\columnwidth]{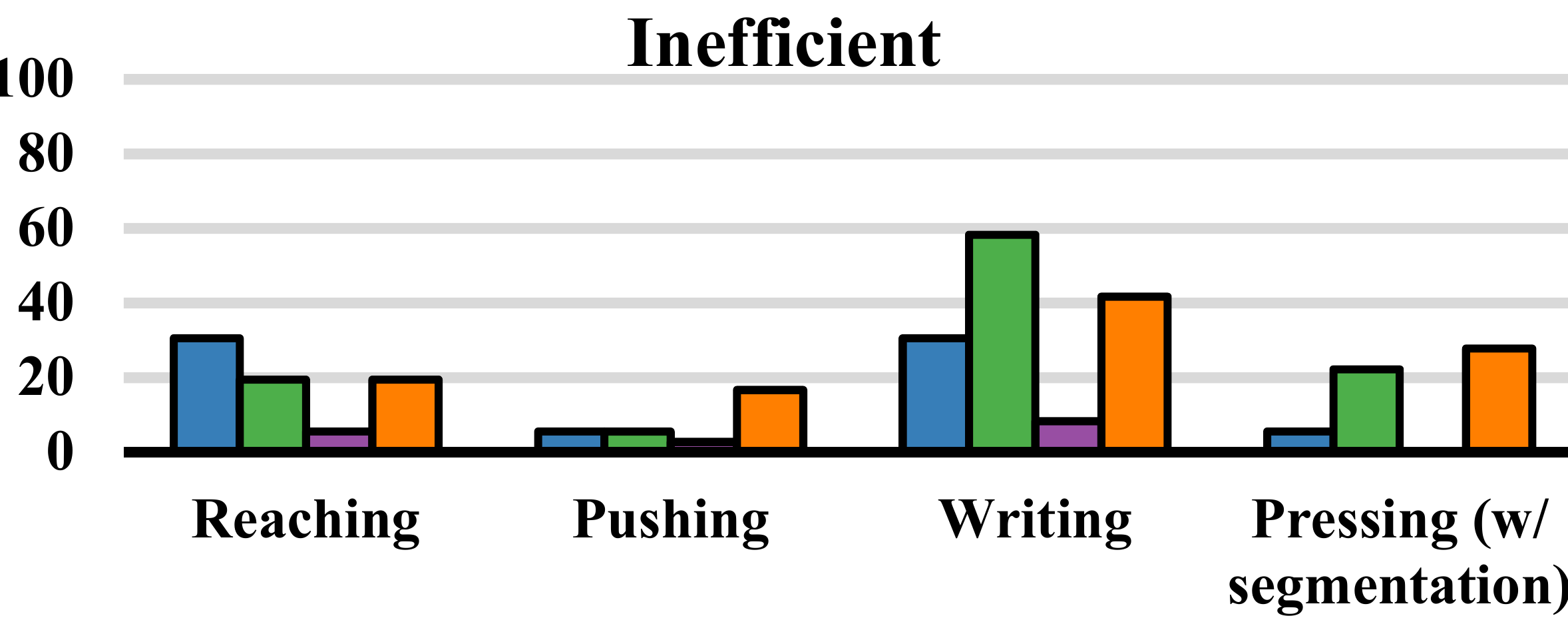}\par
    \includegraphics[trim={2cm 6cm 2cm 6cm}, clip, width=0.95\columnwidth]{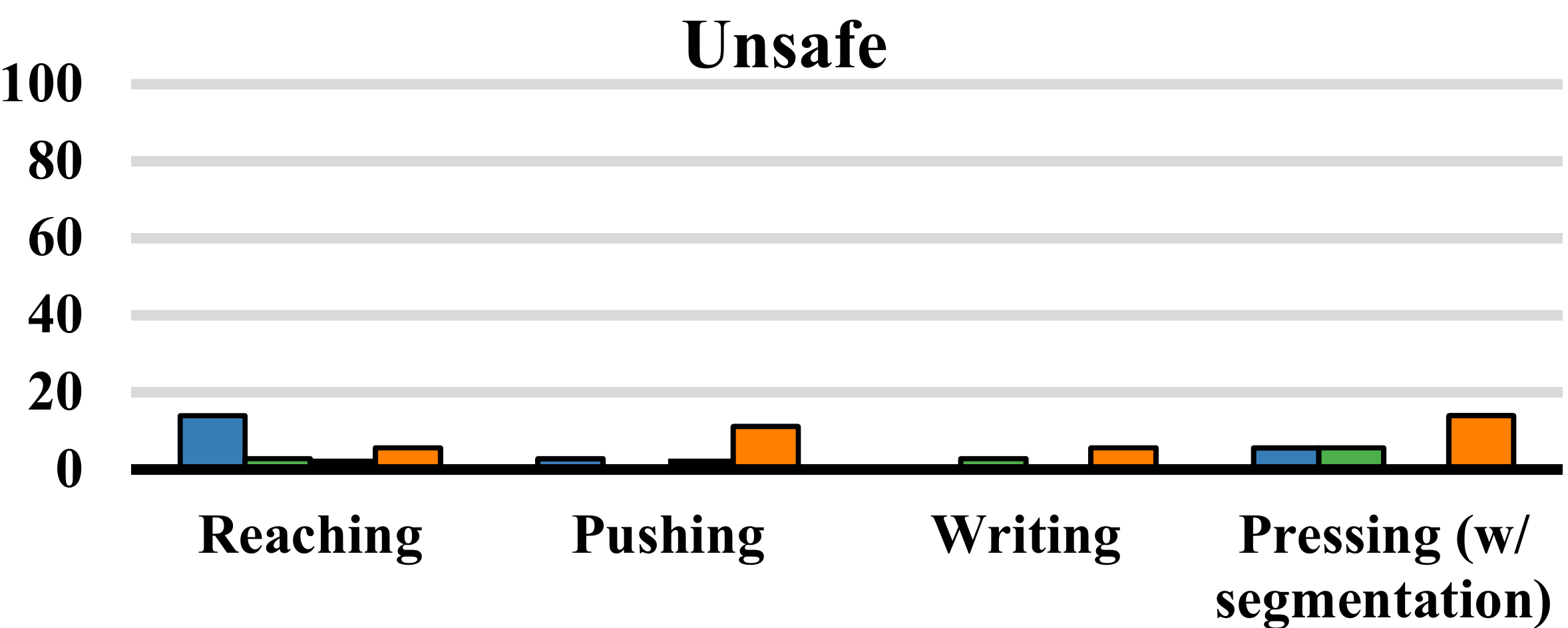}\par
\end{multicols}
\vspace{-0.5cm}
\caption{\small{Subjective user feedback as a percentage of the total number of executions evaluated. Same color code used as previously.}}
\label{fig:feedback}
\end{figure*}

\vspace{-1mm}
\section{Data Processing and Validation Scenarios}\label{sec:Data_Processing}
This section provides an overview of the data processing and parameter tuning methods used in our evaluation, as well as the design of the starting robot configurations used in evaluating the generalization of the chosen approaches.

\vspace{-1mm}
\subsection{Data Preprocessing}\label{sec:Data_Preprocessing}\vspace{-1mm}
Captured human demonstrations consist of robot end-effector poses over time. First, we applied a low-pass moving average filter to remove high-frequency noise from the raw data. Additionally, we estimated the velocities of the end-effector using 1st-order finite differencing. Finally, for methods that require time-aligned trajectories, we also warped the demonstrations to be of the same time duration using dynamic time warping (DTW)~\cite{muller2007dynamic}.

\subsection{Motion Segmentation} \label{sec:motion-seg}
Unlike the other tasks, \emph{pressing} can be seen as two separate tasks or primitives; that is, pressing the first peg followed by pressing the second peg. We assume that considering these two segments as one was likely to adversely affect the performance of some of the approaches. Hence, to ensure fairness in our comparisons, we conducted experiments of  the \emph{pressing} task once \emph{without} segmentation and once \emph{with} segmentation. We performed an additional pre-processing step of motion segmentation~\cite{niekum2015learning,konidaris2012robot, kroemer2015towards, meier2011movement} for the \emph{pressing} task and made a separate dataset for this variation. Specifically, we passed the demonstrated trajectories through a changepoint detection routine~\cite{adams2007bayesian}, which segments the trajectories where peaks are encountered in the normalized velocities. The output was further manually checked to ensure good segmentation. For a given approach, we trained a model per segment, reproduced the task segments separately, and stitched the reproduced segments together to be executed by the robot as one trajectory.  Throughout the paper we clarify which variant of \textit{pressing} is being used, and we evaluate the effect of segmentation on performance in Section \ref{sec:eval-segmentation}.

\vspace{-1mm}
\subsection{Parameter Tuning}\label{sec:Param}\vspace{-1mm}

Our work is motivated by potential real-world applications of motion-based LfD methods, such as factory operation.  To mimic a realistic operational context for the robot, we chose to use only a single common set of parameters for each algorithm.  More specifically, we tuned a parameter set for each \textit{algorithm} for trajectory learning in general, but did not tune unique parameters \textit{per task}, since this would be impractical in real-world settings with novice users. We performed the tuning process on the LASA dataset~\cite{khansari2018lasa} and a small randomly selected subset of robot demonstrations. We manually tuned the parameters of each method until we observed consistently good performance across the test set.
\vspace{-1.5mm}
\subsection{Starting Positions for Generalization}\label{sec:Starting_Pos}
\vspace{-1mm}
Each task model was evaluated from four different initial configurations, $S1$-$S4$, to validate the generalizability of the learned models.  Figure~\ref{fig:task-executions} (bottom row) visualizes the initial positions for each task, overlaid over a set of example demonstrations provided by a participant. S1 was selected to be within $90\%$ confidence interval around the mean of the initial positions of the demonstrations.  S2-S4 were selected outside this range, such that $d(S3)>d(S2)>d(S1)>d(S4)$, where $d(.)$ denotes the distance to the target object. S2 and S3 were chosen to be farther away from the target object, while S4 was chosen to be closer to the object. 

\vspace{-1mm}
\section{Generalization Performance across starting positions and tasks}
\vspace{-1mm}
\label{sec:generalization}
In this section, we study how the average rating for each skill learning method varies across two independent variables: (1) starting position and (2) task. The results are visualized as radar plots in Figs. \ref{fig:starting-pos-s1} through \ref{fig:starting-pos-s4}, where each radar plot reports the average ratings against the executed tasks, for a particular starting position (e.g., S1). The average ratings are computed over nine datapoints corresponding to the nine recorded videos, where each video represents a model query at the given generalization scenario. 
Also reported in Fig. \ref{fig:lineplot-rating}(\emph{top}) are ratings, further averaged against all tasks, per starting position, while Fig. \ref{fig:lineplot-rating}(\emph{bottom}) plots ratings against tasks, averaged against all starting positions. Furthermore, we also provide an analysis of the feedback provided by the evaluators as answers to Q2 in Section \ref{sec:AMT}. Fig. \ref{fig:feedback} reports the feedback -- where the bar charts represent the number of times a particular reason was cited for a given generalization scenario -- as a percentage of the total number of robot executions/videos for that same scenario. 

\subsection{Trends across starting positions}
We see larger variations in average performance of approaches across tasks when the distance between the robot's starting position and the target location is progressively increased (S1 through S3), as shown in Fig. \ref{fig:starting-pos-s1} through \ref{fig:starting-pos-s4}. In general, as evident from Fig. \ref{fig:lineplot-rating}(\emph{top}), we noticed worsening performance with increasing distance. The worsening performance is particularly noticeable for the \emph{writing} and \emph{reaching} tasks in \ref{fig:starting-pos-s2} and \ref{fig:starting-pos-s3}. However, when the distance to the target is significantly decreased, i.e. for starting position S4, CLFDM and TLGC performed consistently in an acceptable manner across the tasks, while ProMP and TpGMM generally under-performed. Overall, TLGC was observed to be least affected by the changes in starting positions for the \emph{pushing}, \emph{writing}, and \emph{pressing} tasks. However, on the \emph{reaching} task, where the other approaches performed generally well, TLGC performed the worst and often at an unacceptable level. 

\subsection{Task-wise evaluation and subjective user feedback}
Analysis in this subsection is based on Fig. \ref{fig:lineplot-rating} (\emph{bottom}) in conjunction with subjective user feedback from Fig. \ref{fig:feedback}. The video accompanying this paper shows some of the failure/success cases mentioned here.

For the \emph{reaching} task, TLGC is hypothesized to have accrued low ratings due to robot executions which often stopped a small distance away from the target. Users often marked these executions as \emph{incomplete}. CLFDM was found to not generalize well for starting positions S2 and S3 which are farther from the target, and had a high percentage incomplete, inefficient, and sometimes unsafe ratings. We hypothesize that this is due to often long and unpredictable paths generated by CLFDM. Furthermore, due to this unpredictability, the robot often collided with the table and hence failed to complete the task, thus the evaluators often marked the executions as \emph{incomplete} and \emph{unsafe}.

On \emph{pushing}, although all approaches on average were consistent across starting positions, we did notice several failure cases. TpGMM was sometimes perceived as \emph{inefficient} and \emph{unsafe} when starting too far away from (S3) or too close to the box (S4). During some of these executions, the robot pushed the lid into the box (farther than the closing point) and dismounted the box. 

For \emph{writing}, only TLGC generalized across starting positions. CLFDM was observed to be the second most consistent across starting positions, except when starting away from the final position (S3). CLFDM often drew a longer L-shaped curve instead of the desired S-shape, which was marked as inefficient and incomplete although it was mostly smooth and safe. Executions by ProMP were frequently marked as incomplete and inefficient since it was often observed to draw non-smooth curves when starting farther away, i.e. S2-S3 or illegible shapes when starting closer (S4). When starting from S3, TpGMM was also found to draw an S-shaped curve with relatively sharp edges. Lastly, for both TpGMM and ProMP, the robot was frequently observed to go back a short distance from S4 before drawing, often penalized by evaluators for being inefficient. 

For the \emph{pressing} task, TpGMM was severely affected by the variations in starting positions. TpGMM was frequently observed to carry out extraneous motions for S3 and S4, often failing to press any of the pegs. Moreover, on a few occasions, TpGMM followed a pressing motion but stayed higher than the height of the pegs. Such executions were often rated incomplete and inefficient. ProMP was sometimes marked inefficient, which can be attributed to jerky and/or extraneous motions when started far away from the pegs.

\subsection{Effect of motion segmentation}
 \label{sec:eval-segmentation}
\begin{wrapfigure}{r}{0.4\columnwidth}
  \begin{center}
        \includegraphics[trim={0cm 0.1cm 0cm 0.15cm}, clip, width=0.35\columnwidth]{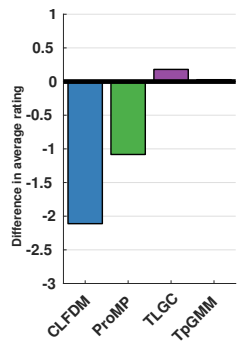}
  \end{center}
\caption{\small{Difference in ratings on the \emph{pressing} task w/ and w/o segmentation.}}
\label{fig:seg-comparison}
\end{wrapfigure}
We conducted an additional evaluation to test our hypothesis regarding the adverse effect of learning on unsegmented data on the  \emph{pressing} task's performance. We trained each algorithm on unsegmented data and performed the same crowdsourced rating in Section \ref{sec:AMT}. Fig. \ref{fig:seg-comparison} shows a bar chart comparing performance with and without segmentation.  Each bar shows the average rating \textit{without} motion segmentation subtracted from the average rating \textit{with} the segmentation routine. We observed that ProMP, and especially CLFDM, suffer significantly when segmentation is not used. This is an expected result for CLFDM, which is known to be incapable of learning self-intersecting motions~\cite{khansari2014learning}. This behavior is in fact expected for all LfD approaches which learn first-order dynamical systems from demonstrations\cite{ravichandar2017learning, neumann2015learning, perrin2016fast, khansari2011learning}. 

\section{Performance Across Experience Level}\label{sec:ratingvsexp}
\begin{figure}
	\centering
	\includegraphics[trim={0cm 0cm 0cm 0cm}, clip, width=1\columnwidth]{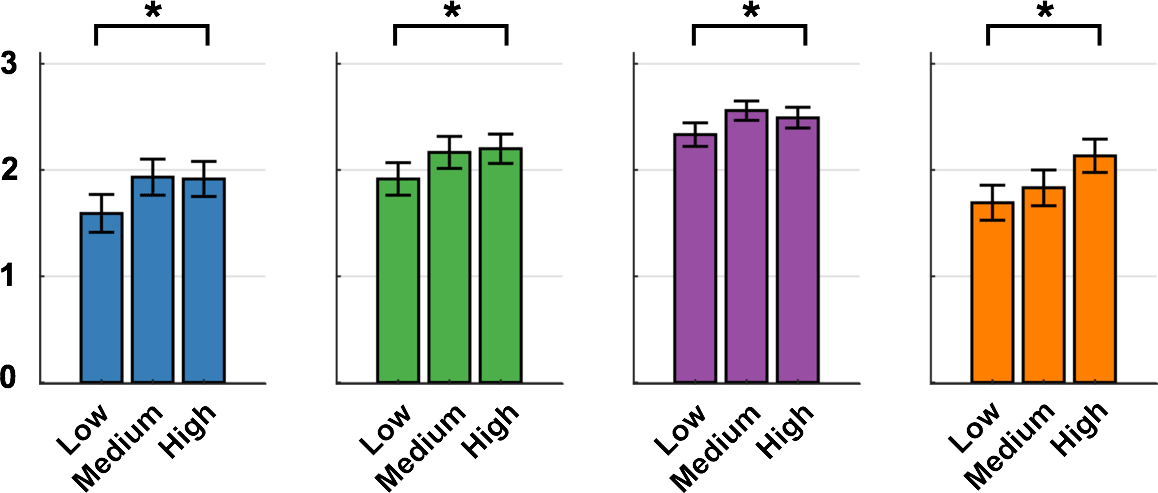}
	\caption{\small{Average ratings grouped by algorithm (CLFDM, ProMP, TLGC, TpGMM) over the experience level of the demonstrators.}}
	\label{fig:exp-level}
\end{figure}

\vspace{-1mm}
In this section, we present an analysis on the dependence of the evaluator ratings, averaged over all the tasks and starting positions, on the experience level of the demonstrators. Fig. \ref{fig:exp-level} provides a visualization of the results.

All the methods show similar increase in average rating from low to high experience when each algorithm is individually observed across experience levels. To corroborate this trend, we also carried out a two-way ANOVA analysis for the approaches against the experience levels. We found that the experience level has a statistically significant effect on average ratings ($p = 0.0389 < 0.05$), while no statistically significant interaction effect was found between the two variables ($p=0.95>0.05$). We further carried out Tukey's range test, which determined that there was a statistically significant effect on performance between the low and high experience levels ($p<0.05$). However, no statistically significant difference in performance was found for low and medium, or medium and high experience levels. A secondary analysis was also carried out on the reasons the evaluators provided for their ratings. This showed that there was a statistically significant difference between user experience levels low and high ($p<0.05$) for a video being marked as inefficient. This means that the evaluators considered the lower-rated videos corresponding to the low experience demonstrators to be more inefficient on average.

In conclusion, we see that higher demonstrator experience positively affects performance across all algorithmic conditions. Interestingly, little difference in performance is observed between participants with high and medium levels of experience (participants with kinesthetic teaching experience vs. participant with general robotics experience).  This observation indicates that having prior knowledge about robots, sensing, or sensitivity to noise is potentially more important than having specific experience with kinesthetic teaching. This insight could direct future work on developing training guidelines to quickly increase novices' expertise. Additionally, an extension may study whether providing supplementary directions (e.g. about speed, waypoints, and direction of motion) to novice users beyond the baseline instruction improves overall performance.

\section{Quantitative Metric Evaluations}

\begin{figure}
  \begin{subfigure}[!htb]{1\linewidth}
    \vspace{2mm}
    \centering
    \includegraphics[trim={0cm 0cm 0cm 0cm}, clip, width=1 \columnwidth, scale=0.93]{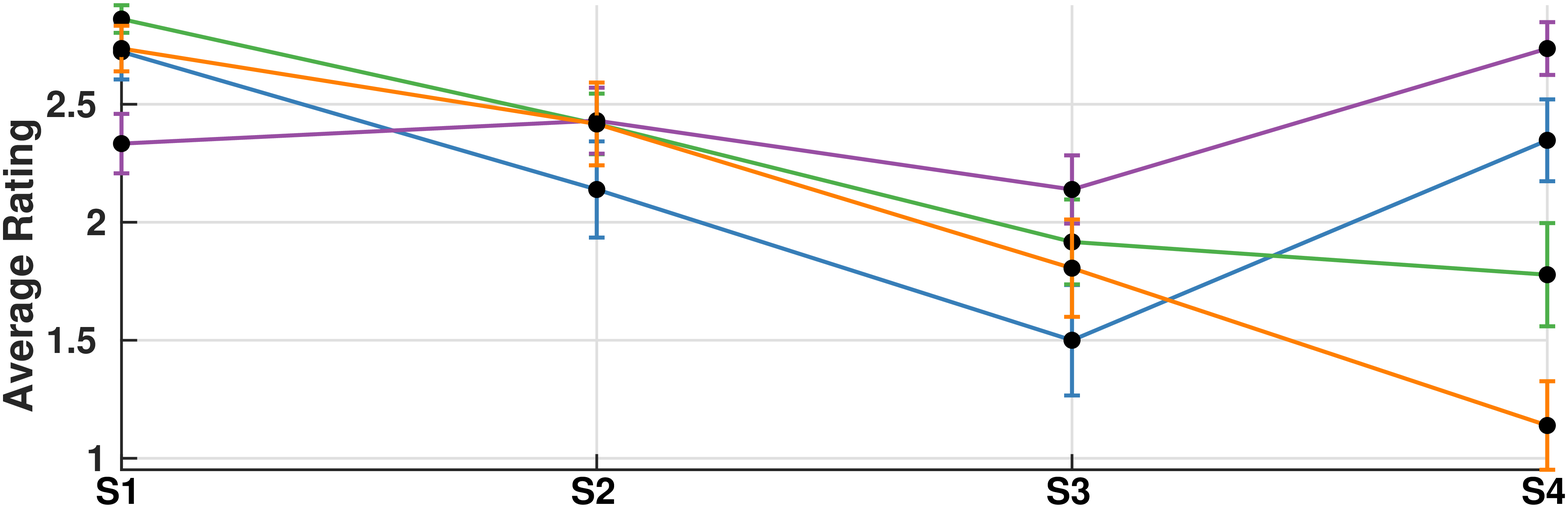}
    \label{fig:rating-vs-starting}
  \end{subfigure}
  \hfill
  \begin{subfigure}[!htb]{1\linewidth}
    \centering
    \includegraphics[trim={0cm 0cm 0cm 0cm}, clip, width=1\columnwidth]{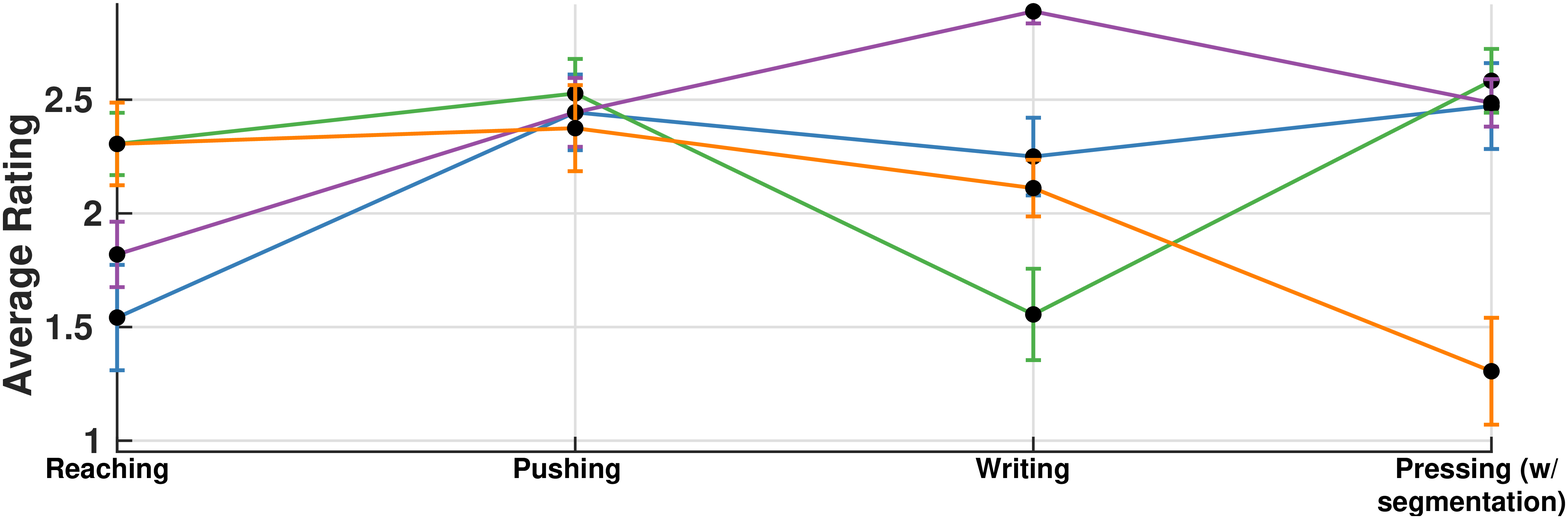}
  \end{subfigure}
  \caption{Plot of average user rating against \emph{top}: starting positions and \emph{bottom}: tasks.}
  \label{fig:lineplot-rating}
\end{figure}

\begin{figure}
	\begin{subfigure}[b]{1\columnwidth}
	    \includegraphics[trim={0cm 0cm 0cm 0cm}, clip, width=1.0\columnwidth]{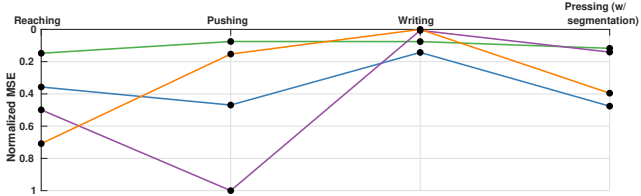}
    \end{subfigure}
	\caption{\small{Normalized mean squared error for different algorithms across all tasks. Note that the vertical axis direction is flipped.}}
	\label{fig:metric-eval}
\end{figure}

Results in the previous sections focused on qualitative measures of performance found using AMT ratings. However, most existing LfD works use quantitative metrics for this purpose, whereby the accuracy of the approach in reproducing the demonstrations themselves is often reported. A widely used metric is the mean squared error (MSE)~\cite{lemme2015open}. We examine whether there is a correlation between the MSE and the ratings we obtained from human evaluators. 

We first reproduced demonstrations by querying the trained skill models from the same initial positions as the demonstrations. To account for the difference in speed between demonstrations and reproductions, we further used dynamic time warping (DTW). The MSE is then given by:
\begin{equation*} 
MSE(\bm{x},\bm{y}) = \frac{1}{N}\frac{1}{T}\sum_{n=1}^{N}\sum_{t=0}^{T}\| \bm{x}_{t,n} - \bm{y}_{t,n} \|^2
\end{equation*}
where $\bm{x}_{t,n}$ and $\bm{y}_{t,n}$ are the datapoints from the demonstrated and time-aligned reproduced trajectories respectively. Furthermore, $T$ is the length of the demonstration while $N$ is the number of demonstrations in the demonstration set. Figure \ref{fig:metric-eval} reports the MSE scores, averaged over starting positions and demonstrators, plotted against the tasks. The vertical axis represents MSE scores, normalized to lie in the range 0 to 1. Note that the direction of the vertical axis for MSE scores has been reversed such that moving up the vertical axis corresponds to improvement in performance in terms of MSE. To compare against the user ratings, we make use of the average user ratings against the tasks plotted in Fig. \ref{fig:lineplot-rating}(\emph{bottom}). For each task, we ranked the approaches in terms of the MSE scores and the user ratings respectively and compared the two rankings. 

Overall, despite a common assumption to the contrary, we observe that MSE is not an accurate predictor of generalization performance of a skill learning approach. This is particularly evident for the \emph{writing} task. For this task, the AMT users were observed to care more about the shape of the executed motion as opposed to its position profile. However, MSE only measures deviations in positions from the demonstrations. Hence, while all the approaches were predicted to perform well according to MSE, only TLGC was able to draw an S-shape curve on most occasions and hence get high ratings.
Furthermore, we also observe that MSE gives little information about the capability of a model to achieve the task goals. In particular, for the \emph{pushing} task, we see that all the approaches were rated highly since they mostly achieved the goal of closing the box towards the end of execution. The users were observed to care less about the trajectory while approaching the box. However, MSE considers the entire length of the trajectories, therefore approaches that fit the data better received higher scores. 

\section{Conclusions and Discussion}
We have presented a large-scale evaluation of four skill learning approaches across four real-world tasks. Our conclusions are based on 720 robot task executions and 3600 ratings provided by AMT users who evaluated the robot trajectories in terms of safety, efficiency, and success in achieving the goals of the task. 

Here, we share algorithm-specific observations to guide users in selecting the appropriate method for their use case.

\subsection{Algorithmic Observations}

For those who plan to use a dynamics-based approach such as CLFDM, it may be useful to note that while such methods guarantee reaching a target location, they cannot guarantee safety or efficiency of executions. However, both these factors have great significance in the real world, as noted by the evaluators who rated CLFDM on the \emph{reaching} and \emph{writing} tasks. CLFDM is also more sensitive than others to changes in distance from the target, but this can be mitigated by segmenting the task, particularly for those with a strong position and direction-of-motion constraint.

Time-parametrized approaches, such as ProMP and TpGMM, can be suitable on tasks which impose minimal direction-of-motion constraint alongside position constraint towards the end (e.g., \emph{pushing} and \emph{reaching}). However, starting very close to the goal can immensely affect performance. This is because time-parametrized approaches are not robust to large spatio-temporal perturbations. Care should be taken to ensure that the robot does not start too close to the final position unless a majority of the provided demonstrations are in the vicinity of this desired starting position. 

For tasks with strong constraints in the direction of motion, a geometric approach like TLGC can be more suitable. We conclude this by observing the consistency of TLGC's performance on the \emph{writing} and \emph{pressing} tasks. This is primarily because TLGC explicitly encodes the shape of the demonstrated motions and minimizes deviations from this shape during reproduction.

\subsection{Research Insights}
This subsection provides general, algorithm-independent research insights learned from this benchmarking effort. We hope this knowledge will guide researchers in developing more robust techniques.

\begin{itemize}[noitemsep,topsep=0pt,parsep=0pt,partopsep=0pt]

\item Approaches with different model representation perform differently over tasks with various constraints. Our evaluations suggest that none of the approaches worked well across all the tasks. While TLGC, the approach with a geometric representation, worked well for tasks with strong constraints in the direction of motion (e.g., \emph{writing}), ProMP, with a time-parametrized probabilistic representation, was found to be most consistent on tasks with positional (e.g., goal location) constraints.

\item Generalization quality decreases as the new starting positions go farther from the original starting positions. None of the approaches were able to consistently generalize to such starting positions in a manner acceptable to the end user.

\item Task complexity affects the generalization capability of the approaches. Our results show that algorithms could generalize better for tasks with simpler constraints and usually struggled over tasks with directional and positional constraints. 

\item For long-horizon tasks with multiple position constraints (e.g. via-points) alongside constraints on the direction of motion, motion segmentation can be a useful pre-processing step to mitigate some of the limitations of skill learning approaches. 

\item Higher user experience level positively impacts the performance of the approaches. Our findings also show that the algorithm performance is affected by the quality of demonstrations provided by users with different levels of experience.

\item Conventional metrics may not be good predictors of approach performance. We have found that the quantitative mean squared error does not serve as a reliable predictor of performance across many tasks. 
\end{itemize}

\addtolength{\textheight}{-12cm} 

\bibliographystyle{IEEEtran}
\bibliography{references}

\end{document}